%% file: medIA/main_arxviz.tex
\documentclass[times,twocolumn,final]{elsarticle}
\usepackage[ruled,vlined]{algorithm2e}
\usepackage{medIA/arxiv}
\usepackage{graphicx}%
\usepackage{multirow}%
\usepackage{amsmath,amssymb,amsfonts}%
\usepackage{amsthm}%
\usepackage{mathrsfs}%
\usepackage[title]{appendix}%
\usepackage[table, xcdraw]{xcolor} 
\usepackage[dvipsnames]{xcolor}
\usepackage{textcomp}%
\usepackage{manyfoot}%
\usepackage{booktabs}%
\usepackage{algpseudocode}%
\usepackage{listings}%
\usepackage{multirow}
\usepackage{pifont}%
\usepackage{adjustbox}
\usepackage{graphicx}
\usepackage{hyperref}
\usepackage{float}
\usepackage{placeins}
\usepackage{fancyhdr}
\usepackage{dsfont}
\usepackage[flushleft]{threeparttable}
\usepackage{color,colortbl}

\definecolor{Gray}{gray}{0.85}
\definecolor{LightCyan}{rgb}{0.88,1,1}
\newcommand{\cmark}{\ding{51}}%
\usepackage{booktabs}

\begin{document}
\fancypagestyle{firstpagestyle}{
    \fancyhf{} 
    \renewcommand{\headrulewidth}{0pt} 
    \fancyhead{} 
    \fancyfoot{} 
    \fancyhead[CO]{\em \fontsize{9pt}{8pt}\selectfont}
}

\fancypagestyle{default}{
    \fancyhf{}
    \fancyhead[R]{\thepage} 
    \renewcommand{\headrulewidth}{0.4pt} 
    \fancyhead[LO]{\em \fontsize{9pt}{8pt}\selectfont}
}



\title{DExTeR: Weakly Semi-Supervised Object Detection with Class and Instance Experts for Medical Imaging}
\author[1,2]{Adrien \snm{Meyer} \fnref{corresp}}
\fntext[corresp]{Corresponding author: \texttt{ameyer1@unistra.fr}}
\author[2,3]{Didier \snm{Mutter}}
\author[1,2]{Nicolas \snm{Padoy}}

\address[1]{University of Strasbourg, CNRS, INSERM, ICube, UMR7357, France}
\address[2]{IHU Strasbourg, Strasbourg, France}
\address[3]{University Hospital of Strasbourg, France}

\received{XXX}
\finalform{XXX}
\accepted{XXX}
\availableonline{XXX}
\communicated{XXX}

\begin{abstract}
Detecting anatomical landmarks in medical imaging is essential for diagnosis and intervention guidance. However, object detection models rely on costly bounding box annotations, limiting scalability. Weakly Semi-Supervised Object Detection (WSSOD) with point annotations proposes annotating each instance with a single point, minimizing annotation time while preserving localization signals. A Point-to-Box teacher model, trained on a small box-labeled subset, converts these point annotations into pseudo-box labels to train a student detector. Yet, medical imagery presents unique challenges, including overlapping anatomy, variable object sizes, and elusive structures, which hinder accurate bounding box inference.
To overcome these challenges, we introduce DExTeR (DETR with Experts), a transformer-based Point-to-Box regressor tailored for medical imaging. Built upon Point-DETR, DExTeR encodes single-point annotations as object queries, refining feature extraction with the proposed class-guided deformable attention, which guides attention sampling using point coordinates and class labels to capture class-specific characteristics. To improve discrimination in complex structures, it introduces CLICK-MoE (CLass, Instance, and Common Knowledge Mixture of Experts), decoupling class and instance representations to reduce confusion among adjacent or overlapping instances. Finally, we implement a multi-point training strategy which promotes prediction consistency across different point placements, improving robustness to annotation variability.
DExTeR achieves state-of-the-art performance across three datasets spanning different medical domains—endoscopy, chest X-rays, and endoscopic ultrasound—highlighting its potential to reduce annotation costs while maintaining high detection accuracy.
\\

\end{abstract}

\maketitle
\thispagestyle{firstpagestyle}

\input{medIA/main/01-introduction}
\input{medIA/main/02-literature}
\input{medIA/main/03-methods}
\input{medIA/main/04-experiments}
\input{medIA/main/05-results}

\input{medIA/main/06-conclusion}
\input{medIA/main/07-acknowledgement}

\bibliographystyle{medIA/sn-basic}
\bibliography{medIA/bib}


\end{document}

%% file: medIA/main/01-introduction.tex
\section{Introduction}

\begin{figure*}[!ht]
\centering
\includegraphics[width=\textwidth]{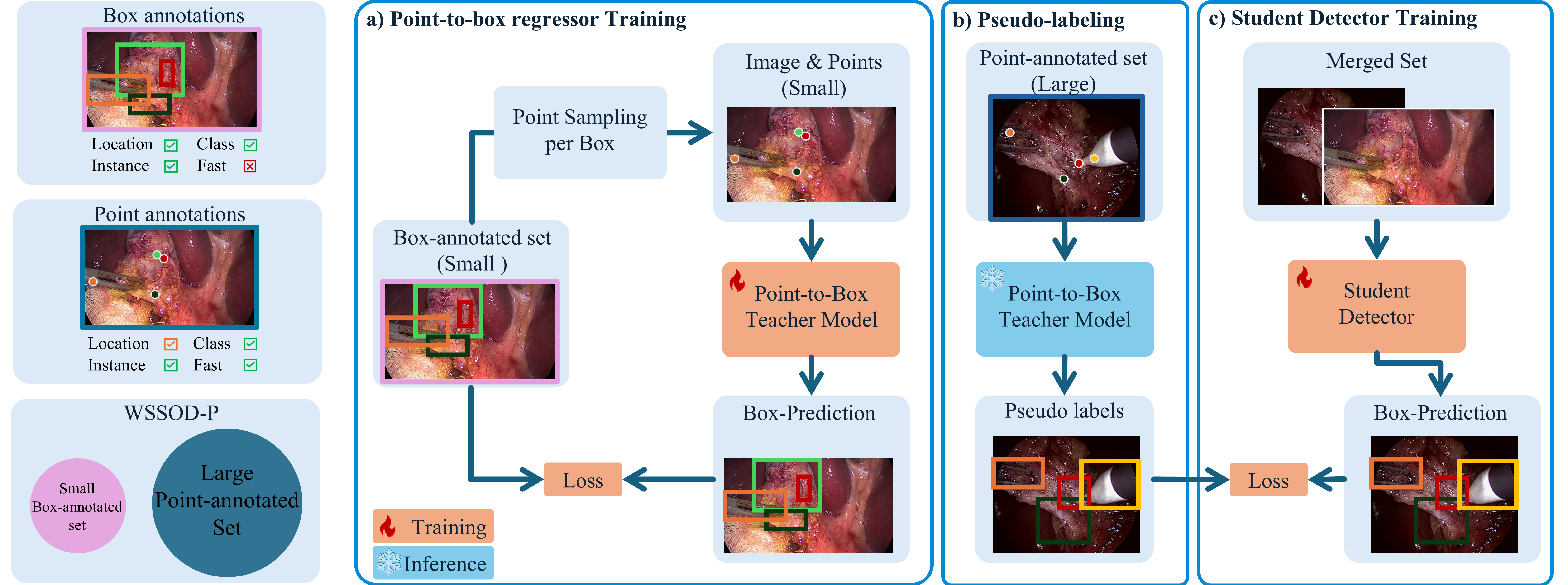}
\caption{Pipeline of the Weakly Semi-Supervised Object Detection with Points (WSSOD-P) framework. The framework aims to train a detector while minimizing box annotation time by predominantly using faster point annotations across the dataset. (a) Box annotations are used to sample points and train a Point-to-Box regressor. (b) The trained Point-to-Box model generates pseudo-box labels for point-annotated images. (c) A student detector is trained on both the box-labeled and pseudo-labeled images to improve detection performance.}
\label{fig:WSSOD_pipeline}
\end{figure*}

Object-detection approaches are becoming essential tools in medical diagnostics, enhancing landmark identification, safety assessments, and anatomy-driven reasoning~\cite{mascagni2022computer,GraphCVS,gbcu}. Deep learning has significantly advanced these technologies; however, its effectiveness relies on the availability of large datasets with detailed annotations~\cite{AIinMed}. In the medical field, creating these datasets is particularly challenging due to the difficulty in annotating complex anatomical structures. These structures often share similar colors and textures, can be highly deformable, and overlap with each other. Consequently, precise annotation is labor-intensive and further compounded by the limited availability of medical professionals.

To mitigate the annotation challenges of medical data and train with limited annotations, Weakly Supervised Object Detection (WSOD) and Semi-Supervised Object Detection (SSOD) approaches have emerged.
WSOD methods use abundant but weakly annotated data, such as image tags~\cite{labonte2023scaling,tang2018pcl}. The studies by~\cite{vardazaryan2018weakly,kim2021weakly} utilize class activation maps to enable detection and localization of surgical tools in endoscopic videos and breast cancer in ultrasound images, respectively, without spatial annotations.
SSOD strategies combine a small set of images with box-level annotations and a larger pool of unlabeled images. These strategies primarily utilize two techniques: First, consistency regularization, which maintains stable detector predictions across differently augmented images \cite{jeong2019consistency}. Second, pseudo-labeling, where a teacher model trained on the labeled dataset generates pseudo-labels for the unlabeled images. A student model learns from both labeled and pseudo-labeled datasets, improving detection capabilities \cite{liu2021unbiased,wang2021data}.
Although both methods lower annotation costs, the effectiveness of the trained detectors often lags behind that of fully-supervised models, depending on the amount of available annotations.

Building on these approaches, Weakly Semi-Supervised Object Detection with Point annotations \cite{pointDETR,PBC,groupRCNN} (WSSOD-P) integrates SSOD and WSOD principles to provide a cost-effective, label-efficient solution for object detection. A Point-to-Box teacher model, trained on a set of box-labeled images, generates pseudo box-labels from point annotated images to train a student model, thus reducing the need for costly box annotation (see Fig. \ref{fig:WSSOD_pipeline}). A single point marks each weakly annotated instance, efficiently conveying both category and location information with a minimal labeling effort, akin to image-level annotations \cite{bearman2016s}.

Prior works~\cite{pointDETR,PBC} introduce a point encoder that elegantly integrates positional encoding and class embeddings to generate point-queries. Following the classic DETR model~\cite{detr}, these point-queries are refined through stacked decoder blocks using self and cross-attention with the feature maps, a feed-forward network (FFN), and a regressor network to predict bounding boxes. However, DETR-based models face notable challenges, including slow convergence and high sensitivity to point locations~\cite{groupRCNN,PBC} as (1) points in the center of instances are often more informative than peripheral points, and (2) adjacent points can share features, blurring instance boundaries and causing confusion.

\begin{figure*}[h!]
\centering
\includegraphics[width=\textwidth]{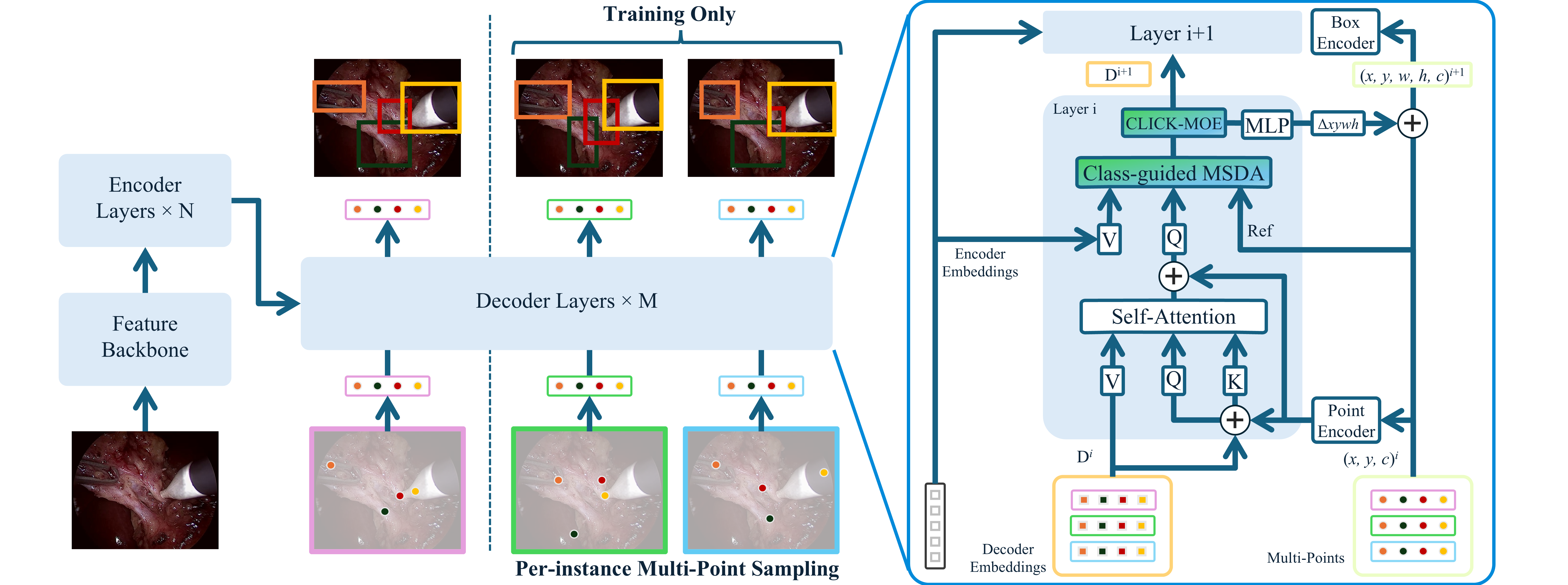}
\caption{DExTeR Model: it leverages multiple point groups during training and incorporates class-guided Multi-Scale Deformable Attention (MSDA) alongside the Class, Instance, and Common Knowledge Mixture of Experts (CLICK-MoE). Point queries are iteratively refined using a Point-to-Box strategy implemented through stacked decoder blocks.}
\label{fig:abstract_visual}
\end{figure*}

To address these challenges, we introduce DExTeR (DETR with Experts), a DETR-based approach within the WSSOD-P framework to enhance object detection in medical imaging. DExTeR analyzes prior information from point-prompts across three key dimensions: the weak \textit{location} information, the strong appearance priors conveyed by \textit{class} information, and the distinct \textit{instance} each point indicates. Effective utilization of \textit{class} and \textit{instance} information is crucial, especially when multiple points are adjacent, as it helps to clarify their relationships and representation. An overview of our method is presented in Fig. \ref{fig:abstract_visual}.

We enhances Multi-Scale Deformable Attention (MSDA)~\cite{defDetr}, which utilizes feature maps at multiple scales from the feature encoder to better distinguish adjacent points. MSDA uses the point location as a reference and derives offsets from the corresponding point-query to pool visual features. By making the point-query class-aware, we guides the attention sampling with class-specific information. We term this approach \textit{class-guided MSDA}.

After undergoing cross-attention to gather visual features, the point-queries are updated through an FFN. However, the FFN operates with fixed parameters for all point-queries, requiring it to handle queries in a class- and instance-agnostic manner—a significant limitation. To better address the specificity of each query and decouple the representation learning process, we introduce our proposed \textit{CLass, Instance, and Common Knowledge Mixture of Experts} (CLICK-MoE). Each query is processed by three experts: the \textit{Common Knowledge Expert}, reminiscent of the FFN used in the transformer block, processes all queries to provide class-agnostic insights; the \textit{Class-specific expert}, which handle queries specific to their respective classes, delivers class-tailored expertise; and the \textit{Instance Expert}, which leverage dynamically generated instance-aware parameters, enhances discrimination among point-queries and reduces confusion between instances.

Finally, we propose a \textit{multi-point training strategy} to reduce point location dependence. In each training iteration, we sample \textit{N} random points per instance and encode each as a point-query. Each point-query interacts with only one other point-query per object, mirroring the inference scenario where a single point per instance is used. Our multi-point training strategy increases the supervision signals and \textit{implicitly promotes consistent predictions} across \textit{diverse point-prompt scenarios} throughout the training process, effectively simulating single-point supervision \textit{N} times per iteration for each instance to enhance training efficiency and robustness.

We validate the effectiveness of DExTeR in data-scarce scenarios across three datasets spanning distinct medical imaging domains: Endoscapes (surgical)~\cite{endoscapes}, VinDr-CXR (X-Ray)~\cite{VinDrCXR}, and EUS-D130 (Endoscopic Ultrasound)~\cite{apeus_dataset}. We summarize our contributions as follows:
\begin{enumerate}
  \item We propose \textit{class-guided MSDA} to direct cross-attention sampling with class-specific information.
  \item We decouple class, instance, and common knowledge processing with our \textit{CLICK-MoE} approach, significantly improving point-query refinement.
  \item We address the dependency on specific point locations with our \textit{multi-point training strategy}.
  \item We demonstrate the effectiveness of our approach across three diverse medical imaging domains for both teacher and student models, achieving consistent improvements in performance under varying annotation settings.
\end{enumerate}

%% file: medIA/main/02-literature.tex
\section{Related Work}

\subsection{DETR-based Supervised Object Detectors}
\label{subsec:detr_based}
DETR proposes a transformer-based end-to-end object detector~\cite{detr} that eliminates the need for many hand-crafted components. It consists of: (a) a backbone that extracts image features from an input image and transforms them into a sequence, with sine/cosine positional encodings added to each feature-point to incorporate spatial context; (b) a multi-layer transformer encoder to refine the features; (c) a set of learnable object queries, used by (d) a multi-layer transformer decoder that employs self-attention for reasoning among object queries and cross-attention for interaction with image features, followed by a feed-forward network (FFN) to enhance representation; and (e) multiple prediction heads. However, DETR suffers from slow convergence and limited feature spatial resolution due to the dense attention modules' limitations in processing image feature maps. Deformable DETR~\cite{defDetr} addresses this by designing a deformable attention module that attends only to certain sampling points around a reference point, accelerating convergence and reducing computational load, thereby allowing multi-scale attention. This reference point concept enables iterative box refinement~\cite{defDetr,dinoDetr}, where bounding boxes are iteratively updated between decoder stages. Another popular technique is DN-DETR's~\cite{dndetr} denoising training method, which speeds up DETR training by feeding noise-added ground-truth queries into the decoder and training the model to reconstruct the original ones.

\subsection{Foundational model}

Recent advances in natural image segmentation have led to the development of foundation models like the Segment Anything Model (SAM)~\cite{SAM}, which yield impressive zero-shot segmentation capabilities from minimal prompts such as points or boxes. However, its adaptation to medical image segmentation, such as with MedSAM~\cite{MedSAM}—a fine-tuned version on medical datasets that only accept box prompt—faces challenges due to the unique characteristics of medical images. Although SAM and MedSAM demonstrate impressive capabilities, they often require multiple prompts and struggle to differentiate highly deformable anatomical structures that share similar colors and textures. Despite the promise of a single foundation model for all modalities, we witnessed the development of domain-adapted foundation model~\cite{zhang2024data}, requiring hugue amount of densely annotated data (i.e., masks). On the one hand, we lack the data and annotations needed to adapt these class-agnostic foundation models to each domain, as they do not leverage prior class information. On the other hand, WSSOD-P models are specifically designed to be trained on small datasets and to effectively leverage class information.

\subsection{Weakly Semi-Supervised Object Detection with Points (WSSOD-P)}
Point DETR~\cite{pointDETR} introduces weakly semi-supervised object detection with point annotations (WSSOD-P). To leverage point annotations, Point-DETR employs a transformer-based Point-to-Box regressor that converts point annotations into bounding-box annotations, enabling the training of an actual object detector. However, it inherits DETR's convergence issues, especially when training data is insufficient~\cite{groupDetr}, which is often the case with WSSOD-P as it aims to train on minimal samples.
Group R-CNN~\cite{groupRCNN} proposes a CNN-based Point-to-Box model to benefit from the better convergence properties of CNNs and leverage multiscale features. They suggest collecting multiple proposals close to the point-prompt to learn an instance-aware representation and improve recall.
Another challenge in WSSOD-P is its sensitivity to the location of point annotations, particularly in medical imaging, where points closer to the center provide more useful information for generating pseudo bounding boxes. PBC~\cite{PBC} introduces point and symmetric consistency strategies to reduce point location sensitivity in X-ray images. Similarly, PSL-Net~\cite{liu2024point} applies point consistency regularization to Group R-CNN. In addition, WSSOD-P can leverage on-the-fly video annotations in medical ultrasound ~\cite{OnTheFlyAnn}.

In contrast, our proposed DExTeR is a transformer-based method that leverages multi-scale features and instance information, achieves fast convergence, and reduces sensitivity to point location.

%% file: medIA/main/03-methods.tex
\section{Methods}

The WSSOD-P pipeline aims to train a Point-to-Box teacher detector to generate pseudo-labels on point-annotated images, which are then used to train a downstream detector (see Fig. \ref{fig:abstract_visual}). The core task is to train an efficient teacher model, and our proposed DExTeR fulfills this role. DExTeR is based on the Point-DETR~\cite{pointDETR} and DETR~\cite{detr} models, following their architecture as detailed in Sec. \ref{subsec:detr_based}.
A ResNet-50 backbone extracts multi-scale visual features from the input image, and a transformer encoder refines these features with MSDA. The points' positions and corresponding classes are encoded into point-queries using a point-prompt encoder. These point-queries first undergo self-attention for inter-query communication, then interact with the image features using \textit{Class-guided MSDA} to extract visual features. The point-queries are further refined with our \textit{CLass, Instance, and Common Knowledge Mixture of Experts} (CLICK-MoE), leveraging the expertise of different experts.
Boxes are then predicted from the refined point-queries. The updated positions, now in the form of boxes, are re-encoded with their corresponding classes to form updated point-queries for further refinement in subsequent decoder blocks. Instead of using a single point-query per instance, we propose a \textit{multi-point training strategy} with \textit{N} groups of point-queries. Each group consists of a different point per instance and does not interact with other groups. We decode each group in parallel, ensuring that the model can predict the corresponding box regardless of the point's position on the instance at any step during training. We detail the novel designs of our architecture in the subsequent sections.

\subsection{Prompt Encoder}
\label{subsec:prompt_encoder}

DExTeR, inspired by Point-DETR, utilizes a prompt encoder to convert point annotations into object queries for the transformer decoder. Unlike DETR's learned positional embeddings, these queries contain precise location and category informations, directly linked to individual object instances.

\begin{figure}[t]
\centering
\includegraphics[width=\columnwidth]{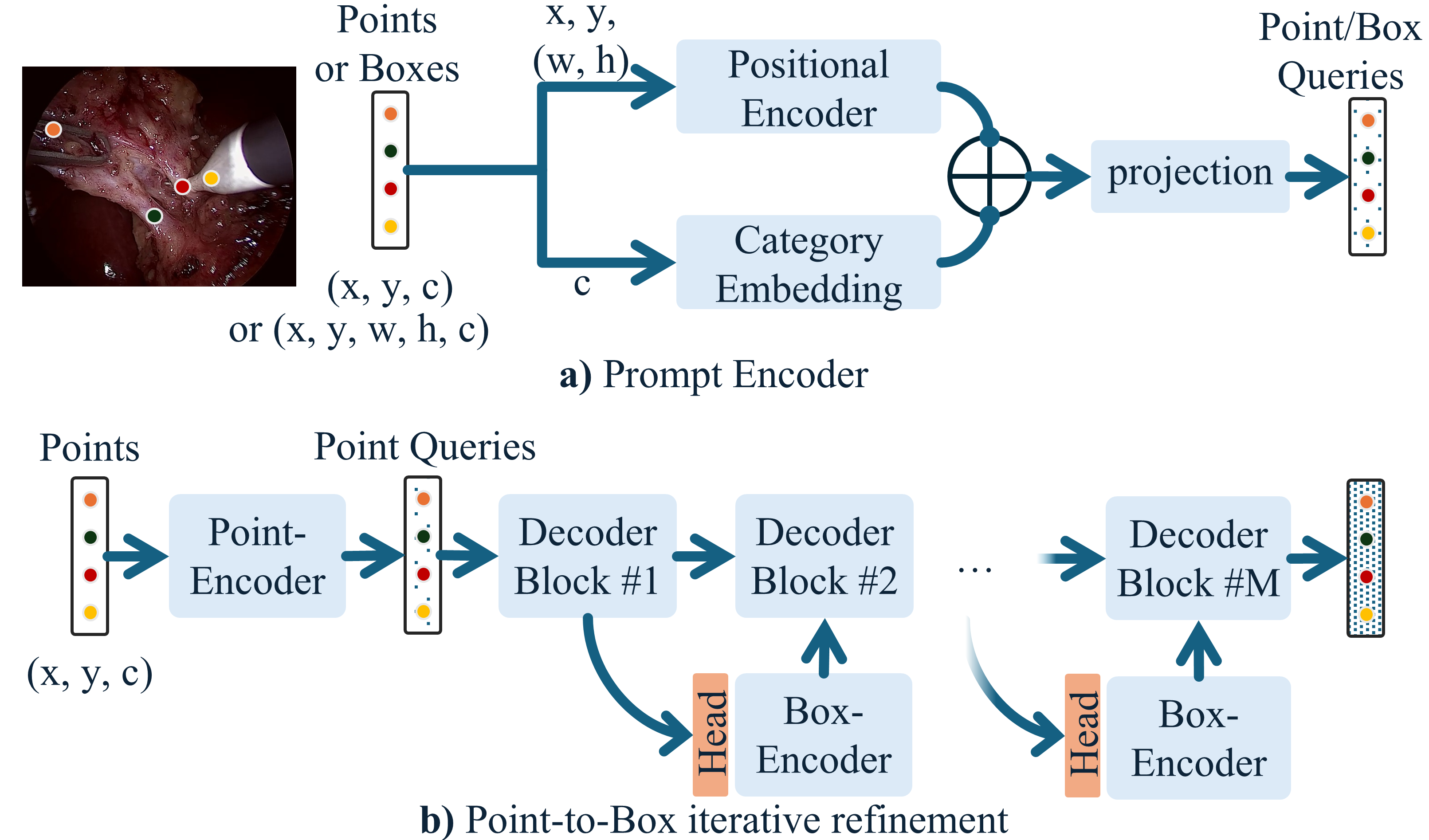}
\caption{Prompt encoding. a) Depending on the input type, either a point encoder or a box encoder is used. Each point \((x, y, c)\) or box \((x, y, w, h, c)\) is encoded using positional encoding and category embedding, then merged by a linear layer into the final representation. b) Point-to-Box refinement strategy: Encodes point annotations and predicts refinements after each decoder stage to form box-queries, which are re-encoded for subsequent refinement.}
\label{fig:prompt_encoder}
\end{figure}

We further introduce a \textit{Point-to-Box iterative refinement} strategy —an adaption from the iterative box-refinement~\cite{defDetr, dinoDetr} methods— which dynamically refines the size and position of bounding boxes throughout the decoding process, thereby enhancing the accuracy and granularity of object detection. After each decoder block, a prediction head generates adjustments (deltas) to transform prompt queries into bounding boxes. The system employs two types of prompt encoders: a point encoder and a box encoder (See Fig.~\ref{fig:prompt_encoder}).

The point encoder operates before the first decoder block, processing a point annotation \((x, y, c)\) into a normalized 2D coordinate \((x, y)\) within \([0,1]^2\) and a category index \(c\). A fixed spatial encoding generates the position embedding \(e_{\text{pos\_xy}} \in \mathbb{R}^{256}\).
In contrast, the box encoder, used between decoder blocks, mirrors the point encoder's functionality but expands its role to encode not only coordinates \((x, y)\) but also dimensions \((w, h)\) (widht and height of a bounding box), resulting in position embedding \(e_{\text{pos\_xywh}} \in \mathbb{R}^{512}\).
Both encoders share a class embedding matrix, dimensioned as \(C \times 256\) (where \(C\) is the number of classes), optimized alongside the model, from which the category embedding \(e_{\text{cat}} \in \mathbb{R}^{256}\) is derived using index \(c\). We concatenate the \(e_{\text{pos}}\) and \(e_{\text{cat}}\) embeddings and use a linear layer to project them into a fixed embedding dimension, forming the final prompt-query embedding. For ease of discussion, we refer to both point and box queries as "point-query."

\subsection{Class-guided MSDA}

Previous implementations of DETR-based frameworks as Point-to-Box regressors have relied on dense cross-attention between point queries and image features. However, the high computational demand of dense attention restricts interactions to only the last feature map, which is of lower resolution. This restriction not only diminishes the point-query location information but also hamperes the model's ability to distinguish between close or overlapping instances.

To leverage multi-scale information, we utilize Multi-Scale Deformable Attention (MSDA)~\cite{defDetr} in both encoder self-attention and decoder cross-attention processes. This method employs feature maps from various scales produced by the feature encoder, thereby enhancing the distinction between adjacent points. MSDA assigns to each object-query a reference point, which is predicted from the encoder output, and calculates offsets and weights for sampling features relative to these reference points.

Building on this, we introduce \textit{Class-guided MSDA}, a refinement where the point-query's own coordinates and class embedding serve as the reference. This modification allows the offsets and weights for feature sampling to be directly derived from the point-query and its coordinates, utilizing the rich class-specific information (such as size and shape) to guide attention sampling more effectively.

This contribution already significantly enhances the basic Point-DETR model through the integration of Class-guided MSDA, the newly proposed Prompt-encoder, and an iterative Point-to-Box refinement strategy (See Section~\ref{subsec:prompt_encoder}), collectively referred to as \textit{Deformable Point-DETR}. Despite these advancements, the model still primarily relies on point location and uses instance and class information only for sampling visual features, not for refining point-queries. To address this limitation, we introduce \textit{CLICK-MoE} to refine point-queries by decoupling class, instance, and common knowledge.


\subsection{CLICK-MoE}



\begin{figure}[t]
\centering
\includegraphics[width=\columnwidth]{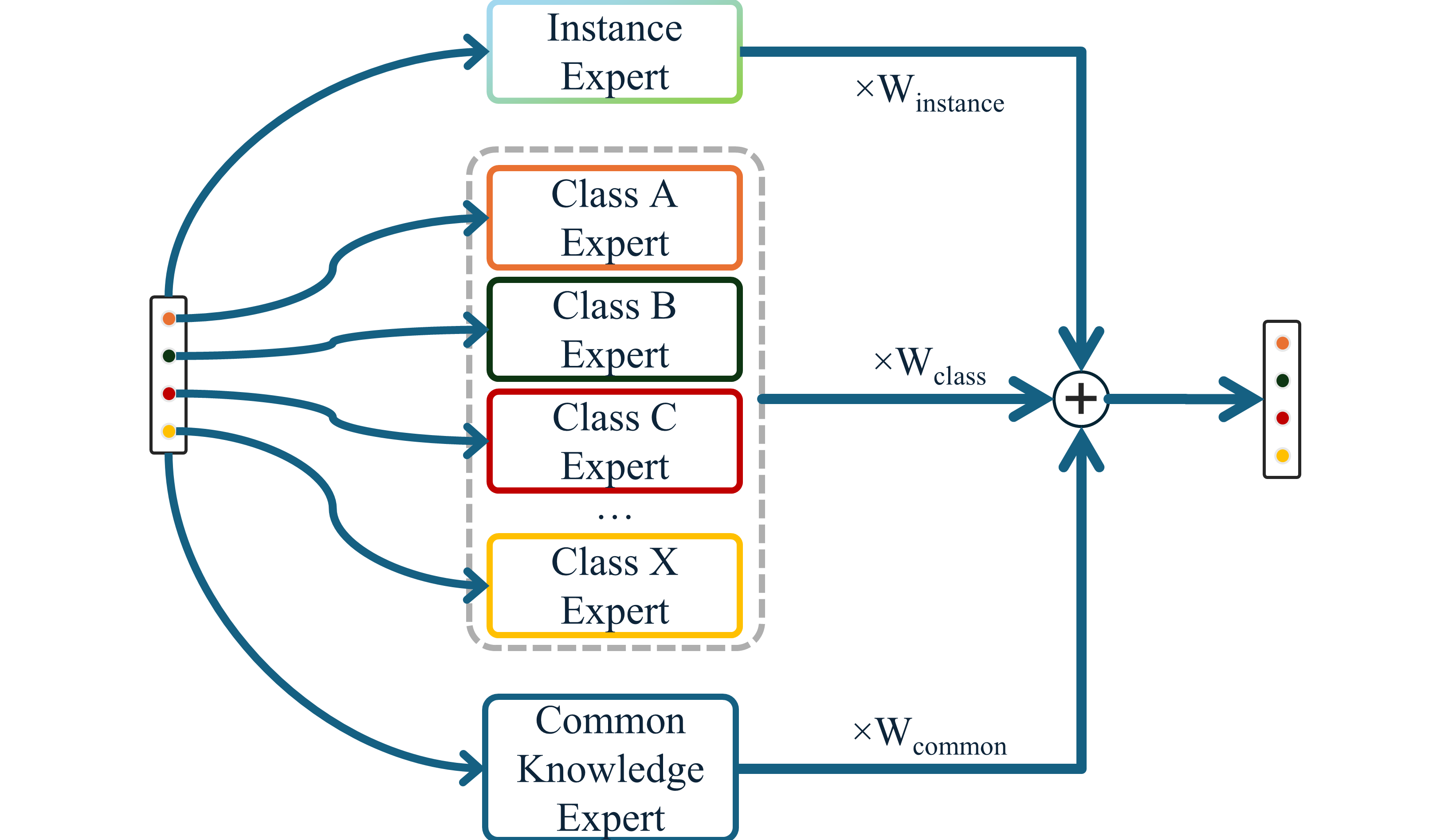}
\caption{Class, Instance, and Common Knowledge Mixture of Experts (CLICK-MoE): input object queries are processed through three specialized experts—a dynamically generated instance expert, a class-specific expert based on the query's label, and a common knowledge expert. The outputs from these experts are combined using weighted element-wise summation, where the weights are \(W_{\text{instance}}\), \(W_{\text{class}}\), and \(W_{\text{gen}}\).
}
\label{fig:click_moe}
\end{figure}

We introduce the \textit{CLass, Instance, and Common Knowledge Mixture of Experts} (CLICK-MoE) to refine point-queries within our transformer-based model (see Fig.~\ref{fig:click_moe}). This layer operates independently for each point-query, replacing the traditional feed-forward network (FFN) sub-block within the decoder block. By employing a decoupled approach with three specialized experts, CLICK-MoE effectively processes and differentiates point-queries based on class, instance, and common knowledge, offering a flexible architecture that could potentially adapt to various downstream datasets.

\begin{figure}[t]
\centering
\includegraphics[width=\columnwidth]{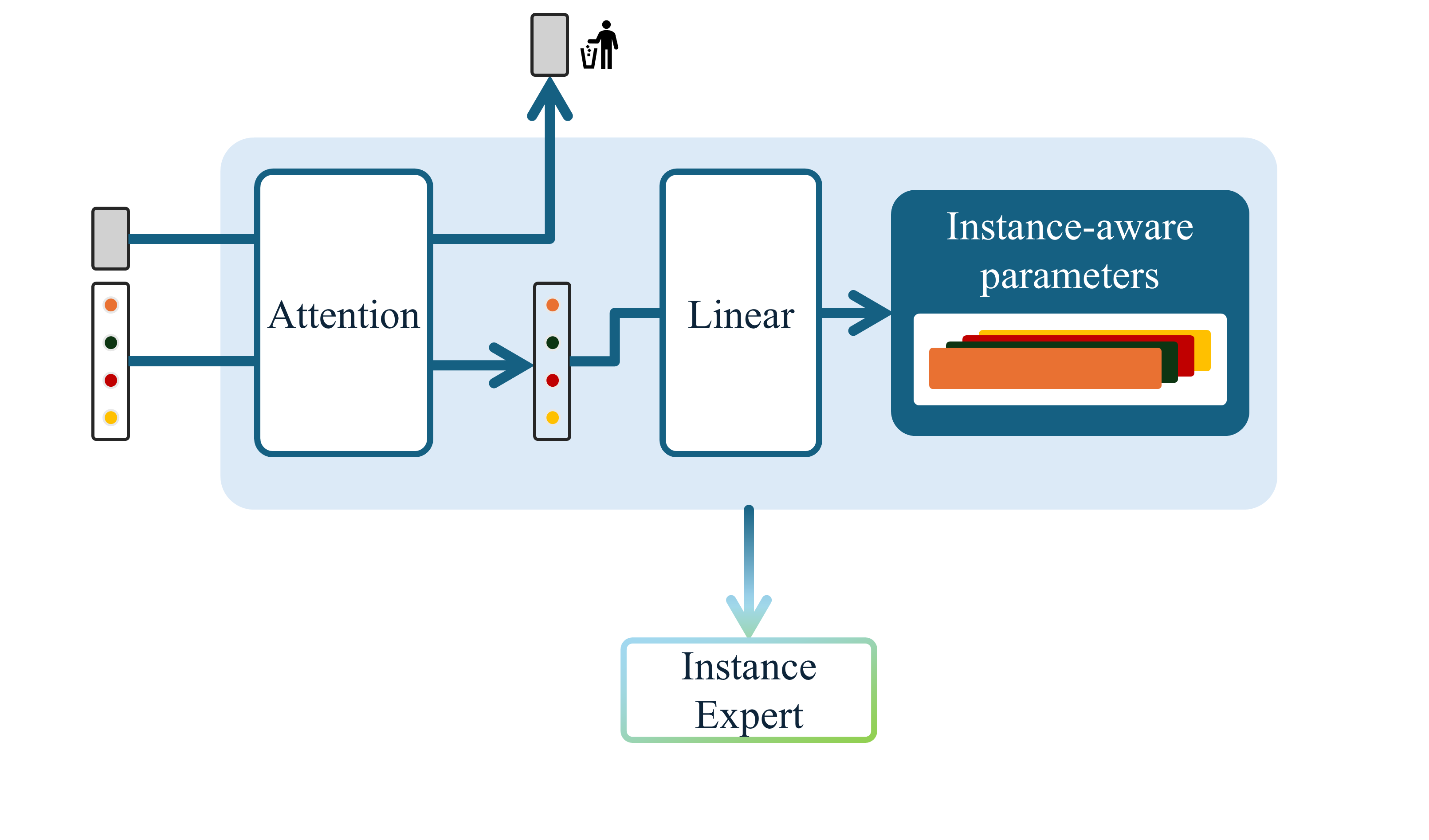}
\caption{Instance Expert weight generation. The object queries, along with a learnable embedding, undergo self-attention and are projected with a linear layer to produce instance-aware parameters for the instance experts.
}
\label{fig:weight_gen}
\end{figure}

\begin{itemize}
    \item The \textit{Common Knowledge Expert}, a two-layer FFN with ReLU activation after the first layer, processes all queries to provide general, class-agnostic insights. The Common Knowledge Expert's generalized insights are directly transferable for fine-tuning in transfer learning scenarios.
    \item The \textit{Class-specific Experts}, similar in structure to the Common Knowledge Expert, only process queries specific to their respective classes to deliver class-specific knowledge. Given that most datasets have a manageable number of classes, one expert per class is feasible. In transfer learning scenarios, these experts are replaced with new ones to accommodate different classes, without compromising the other experts.
    \item The \textit{Instance Experts}, consisting of a single linear layer, use dynamically generated, instance-aware parameters to enhance differentiation between point-queries and minimize confusion between instances. These parameters are derived from the point-queries and a learnable embedding through self-attention, followed by a linear layer (See Fig.~\ref{fig:weight_gen}). The learnable embedding can be seen as the base weights for the instance expert. Through its interaction with the queries during the attention mechanism, it aids in adapting the representation of the point-queries. It is discarded after completing the attention process with the point-queries.
\end{itemize}

The outputs from all three experts is weighted by \( W_{\text{instance}} = W_{\text{class}} = W_{\text{gen}} = \tfrac{1}{3} \),  to promote inter-experts compatibility and coherence, and then summed with the original point-queries. In transfer learning, the Common Knowledge Expert, being directly transferable, serves as a stable base that supports the learning of new class-specific experts, which is particularly beneficial in scenarios involving smaller datasets. We refer to the model incorporating this component as \textit{CLICK-DETR}.

\subsection{Multi-point Training Strategy}


One limitation of WSSOD-P is its reliance on location priors, where point annotations near the center of an anatomical area tend to be more informative than those near the boundaries. This difference in semantic information makes central points more valuable for generating pseudo bounding boxes~\cite{PBC}. Thus, a key challenge is to accurately map point annotations to bounding boxes while accounting for the variability in the annotated points' locations.

We propose our \textit{Multi-point training strategy}. Point-to-Box models are trained using points sampled from box-annotated instances, allowing for infinite point sampling at no additional annotation cost beyond the initial box annotations. Instead of using a single point-query per instance, we sample \textit{N} groups of point-queries. Each group consists of a different point per instance and, through group-wise self-attention, operates independently without interacting with other groups. 
During each training iteration, each group is decoded in parallel, implicitly encouraging the network to consistently predict the correct bounding box for diverse points on a given instance. Our method also enhances the supervision signal, further accelerating training convergence and improving model accuracy. We normalize the loss by the total number of point-queries, calculated as the number of instances multiplied by the number of groups.

Our strategy is employed solely during the training process, effectively simulating the use of a single point per instance \textit{N} times in one iteration to enhance training speed and robustness. At inference time, the model operates with just one group per instance—the annotated points. Since the groups do not interact with each other during training, this setup is consistent with the task the model is trained on.
Our approach can be integrated into any DETR-based Point-to-Box model, accelerating training convergence and improving the model’s ability to accurately predict bounding boxes.

%% file: medIA/main/04-experiments.tex
\section{Experiments}

\subsection{Dataset}

Transformer models are difficult to train from scratch and benefit greatly from pretraining. Since Group R-CNN is pretrained on the COCO dataset \cite{msCOCO}, we also pretrain DExTeR and the other baselines on COCO. This pretraining substantially improves performance across all models, and all results reported in this paper are finetuned from these pretrained weights. During training, Point-to-Box regressors are optimized by randomly sampling points within instance bounding boxes and using these boxes as regression targets. At inference, to ensure fair comparison across methods, we generate a single fixed point annotation per instance from the available mask (or bounding box if unavailable). This fixed point is consistently used for evaluation across all methods in weakly annotated datasets. Performance is reported in terms of mean Average Precision (mAP), which measures precision across multiple Intersection over Union (IoU) thresholds (0.5 to 0.95), and mAP@50, which assesses precision at a fixed IoU threshold of 50\%.
We conducted experiments using three datasets from different medical areas: Endoscapes~\cite{endoscapes} (surgical), VinDr-CXR~\cite{VinDrCXR} (X-Ray), and our private EUS-D130 dataset~\cite{apeus_dataset} (endoscopic ultrasound).

\subsubsection{Endoscapes}
Endoscapes contains 201 annotated videos from laparoscopic cholecystectomy dissections. It includes 1933 frames, annotated every 30 seconds, with bounding boxes across six classes (five anatomical and one tool class). We use the official training, validation, and testing splits from~\cite{endoscapes}.

\subsubsection{VinDr-CXR} VinDr-CXR contains 15,000 chest X-ray images annotated for 14 thoracic diseases, such as aortic enlargement and cardiomegaly. Since the VinDr-CXR dataset exhibits a long-tailed distribution with some abnormal categories containing fewer than ten samples, we follow~\cite{PBC} and combined the eight smallest categories into one class.

\subsubsection{EUS-D130} EUS-D130 (private) extends EUS-D50~\cite{apeus_dataset}, comprising 130 videos acquired using curved linear and zoomed radial probes from Olympus and Hitachi EUS systems. Elastography and contrast modes are excluded, with all videos captured at a frequency of 7.5 MHz. Pancreas and lesion annotations are annotated using the MOSaiC platform~\cite{mosaic}.

\begin{table*}
\begin{threeparttable}
\scriptsize
\caption{Results of Point-to-Box teacher models in mAP ($\%$) using 12.5\% and 50\% of the training set. DExTeR significantly improves upon the base point-DETR and outperforms all baselines on the three datasets.}
\label{tab:modernizing_point_detr2}

    \setlength{\tabcolsep}{2pt}
        \begin{tabular*}{\textwidth}{@{\extracolsep{\fill}}ccccccccccccc}
        \toprule
        & \multicolumn{4}{c}{Endoscapes} & \multicolumn{4}{c}{VinDr-CXR} & \multicolumn{4}{c}{EUS-D130} \\
        \cmidrule(lr){2-5} 
        \cmidrule(lr){6-9} 
        \cmidrule(lr){10-13} 
        Models & \multicolumn{2}{c}{$12.5\%$} & \multicolumn{2}{c}{$50.0\%$}
        & \multicolumn{2}{c}{$12.5\%$} & \multicolumn{2}{c}{$50.0\%$}
        & \multicolumn{2}{c}{$12.5\%$} & \multicolumn{2}{c}{$50.0\%$}\\
        \cmidrule(lr){2-3} \cmidrule(lr){4-5}
        \cmidrule(lr){6-7} \cmidrule(lr){8-9}
        \cmidrule(lr){10-11} \cmidrule(lr){12-13}
        & mAP & mAP@50 & mAP & mAP@50
        & mAP & mAP@50 & mAP & mAP@50 
        & mAP & mAP@50 & mAP & mAP@50 \\
        \bottomrule
        \rowcolor{LightCyan}
        \multicolumn{13}{c}{existing sota} \\
        Group-RCNN
        & $27.1{\scriptstyle \pm3.1}$ & $50.7{\scriptstyle \pm2.8}$ & $33.7{\scriptstyle \pm2.4}$ & $61.6{\scriptstyle \pm2.2}$
        & $22.7{\scriptstyle \pm4.4}$ & $51.7{\scriptstyle \pm4.2}$ & $25.4{\scriptstyle \pm2.2}$ & $56.9{\scriptstyle \pm1.9}$
        & $5.9{\scriptstyle \pm2.6}$ & $27.2{\scriptstyle \pm2.4}$ & $8.1{\scriptstyle \pm1.9}$ & $34.5{\scriptstyle \pm1.7}$\\
        
        PBC
        & $17.7{\scriptstyle \pm3.7}$ & $35.4{\scriptstyle \pm3.5}$ & $23.5{\scriptstyle \pm2.5}$ & $49.2{\scriptstyle \pm2.3}$
        & $17.4{\scriptstyle \pm4.8}$ & $38.2{\scriptstyle \pm4.5}$ & $20.7{\scriptstyle \pm2.9}$ & $50.1{\scriptstyle \pm3.0}$
        & $6.2{\scriptstyle \pm3.2}$ & $27.7{\scriptstyle \pm3.5}$ & $8.6{\scriptstyle \pm2.7}$ & $35.9{\scriptstyle \pm2.0}$\\
        
        Point-DETR
        & $16.2{\scriptstyle \pm3.3}$ & $36.0{\scriptstyle \pm3.1}$ & $23.4{\scriptstyle \pm2.7}$ & $48.9{\scriptstyle \pm2.7}$
        & $15.2{\scriptstyle \pm4.8}$ & $38.8{\scriptstyle \pm4.9}$ & $18.8{\scriptstyle \pm3.2}$ & $47.7{\scriptstyle \pm3.0}$
        & $5.9{\scriptstyle \pm3.3}$ & $26.2{\scriptstyle \pm3.1}$ & $8.4{\scriptstyle \pm2.4}$ & $35.4{\scriptstyle \pm2.0}$\\
        
        \bottomrule
        \rowcolor{LightCyan}
        \multicolumn{13}{c}{ours} \\
        Def. Point-DETR
        & $24.0{\scriptstyle \pm3.2}$ & $43.6{\scriptstyle \pm3.0}$ & $31.4{\scriptstyle \pm2.5}$ & $54.3{\scriptstyle \pm2.4}$
        & $21.0{\scriptstyle \pm4.8}$ & $48.8{\scriptstyle \pm4.5}$ & $24.3{\scriptstyle \pm2.5}$ & $54.8{\scriptstyle \pm2.4}$
        & $6.6{\scriptstyle \pm3.1}$ & $28.4{\scriptstyle \pm2.8}$ & $9.0{\scriptstyle \pm2.3}$ & $36.5{\scriptstyle \pm1.7}$\\
        CLICK-DETR
        & $27.8{\scriptstyle \pm2.9}$ & $47.8{\scriptstyle \pm2.9}$ & $33.9{\scriptstyle \pm2.4}$ & $57.4{\scriptstyle \pm2.3}$
        & $21.5{\scriptstyle \pm4.9}$ & $49.9{\scriptstyle \pm4.1}$ & $25.6{\scriptstyle \pm2.1}$ & $56.6{\scriptstyle \pm2.3}$
        & $6.8{\scriptstyle \pm2.9}$ & $29.3{\scriptstyle \pm2.6}$ & $9.4{\scriptstyle \pm2.3}$ & $37.1{\scriptstyle \pm1.4}$\\
        DExTeR
        & $\textbf{31.4}{\scriptstyle \pm2.9}$ & $\textbf{52.8}{\scriptstyle \pm2.7}$ & $\textbf{36.8}{\scriptstyle \pm1.8}$ & $\textbf{62.9}{\scriptstyle \pm2.0}$
        & $\textbf{23.2}{\scriptstyle \pm4.6}$ & $\textbf{52.9}{\scriptstyle \pm4.1}$ & $\textbf{26.5}{\scriptstyle \pm2.2}$ & $\textbf{58.3}{\scriptstyle \pm2.1}$
        & $\textbf{8.1}{\scriptstyle \pm2.7}$ & $\textbf{33.0}{\scriptstyle \pm2.3}$ & $\textbf{11.0}{\scriptstyle \pm2.4}$ & $\textbf{41.1}{\scriptstyle \pm1.5}$\\
        
        \bottomrule
        \end{tabular*} 

\end{threeparttable}
\end{table*}

\begin{figure*}[!h]
\centering
\includegraphics[width=\textwidth]{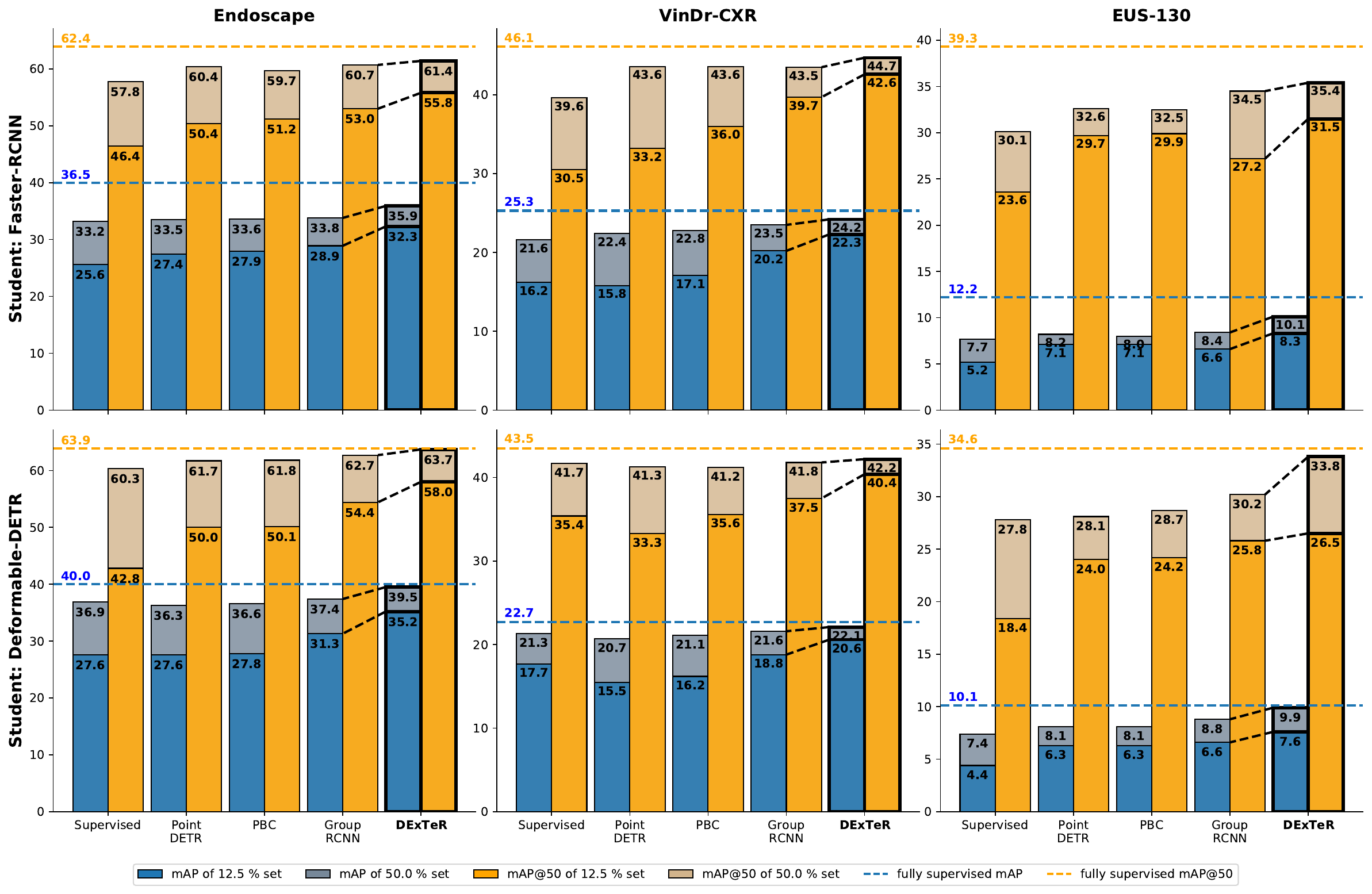}
\caption{Student models' results. Solid blue and light blue bars represent the mAP achieved with 12.5\% and 50\% of the data box-labeled, respectively. Similarly, solid orange and light orange bars represent the mAP@50 under the same 12.5\% and 50\% box-labeled settings. The solid horizontal lines indicate the fully supervised performance for reference. The "Supervised" method refers to a model trained solely on available box-labeled data, without incorporating any pseudo-labels. DExTeR consistently outperforms other state-of-the-art models. Notably, when trained with 50\% box-labeled data and 50\% pseudo-labeled data, the student models achieve performance comparable to the fully supervised baseline.}
\label{fig:student_endo_def}
\end{figure*}

We use only $12.5\%$ and $50.0\%$ of the training data from each respective dataset, simulating a scenario where only a small portion of the data is box-annotated. The remaining training data were used as point-annotated data, pseudo-labeled using the corresponding Point-to-Box model. The student models were then trained on both sets.





\subsection{Implementation Details}

DExTeR was pretrained on COCO using eight V100 GPUs and subsequently finetuned on a single V100 GPU with a batch size of two images per GPU. We followed DETR's data augmentation protocols to maintain consistency. The implementation leverages the MMDetection v3.3 library~\cite{mmdetection}. The model was trained for 24 epochs during both pretraining and finetuning, with the learning rate reduced by a factor of 10 at the 20th epoch. We adopted the AdamW optimizer for network optimization, starting with an initial learning rate of $1 \times 10^{-4}$ and a warm-up period of 500 iterations.  Following deformable-detr~\cite{defDetr}, we use
a linear combination of L1 and GIoU~\cite{giouLoss} loss (5:2).

%% file: medIA/main/05-results.tex
\section{Results}

We first evaluated the performance of the developed Point-to-Box teacher models across datasets. Next, we assessed DExTeR by training student detectors and conducted extensive ablation studies to analyze its proposed components. All experiments were performed using three splits, and the mean results are reported.

\subsection{Point-to-Box teacher models}

We present the quantitative results of the Point-to-Box teacher models in Table~\ref{tab:modernizing_point_detr2}. Results are presented for three models developed iteratively, incorporating the different proposed components, with the final iteration represented by our proposed model, DExTeR. For comparison, we include three state-of-the-art (SOTA) Point-to-Box models: Group-RCNN~\cite{groupRCNN}, PBC~\cite{PBC}, and Point-DETR~\cite{pointDETR}, the latter serving as the foundation for our approach.

The proposed models show significant improvements over the baseline Point-DETR and the best-performing baseline, Group-RCNN. In the 12.5\% box-annotated setting, Deformable Point-DETR achieves an average gain of 4.8 mAP over Point-DETR, while DExTeR further improves this to 8.5 mAP. Additionally, DExTeR surpasses Group-RCNN by 2.3 mAP in this setting. In the 50\% box-annotated setting, DExTeR continues to demonstrate its effectiveness, achieving an average gain of 2.4 mAP over Group-RCNN.

We present qualitative results in Fig.~\ref{fig:student_endo_def} for the 12.5\% setting. DExTeR demonstrates superior performance, producing tighter boxes that better fit objects (rows 1 and 2). In contrast, Group-RCNN often generates overly wide boxes and struggles to distinguish close instances (row 2). DExTeR effectively leverages prior class and instance information, addressing these issues. However, row 3 highlights a failure case where the predicted box for the pancreas does not fully encompass the organ.

\subsection{Student detectors}
We present the results of the student detectors in Fig.~\ref{fig:student_endo_def}, utilizing Faster R-CNN~\cite{fasterRCNN} and Deformable DETR~\cite{defDetr} as student models. The experiments are conducted with either 12.5\% or 50\% of the dataset box-labeled, while the remaining data is pseudo-labeled using the corresponding teacher model under evaluation. We report both mAP and mAP@50 metrics. For both student models, DExTeR, as the teacher model, consistently outperforms all other baselines. Specifically, when using DExTeR, we achieve on average mAP@50 scores that are 86.9\% and 96.6\% of the fully supervised performance, respectively. In comparison, the supervised baseline achieves 67.1\% and 87.6\% of the fully supervised performance, while Group-RCNN achieves 81.0\% and 93.5\% under the same settings.
These results demonstrate that the improvements achieved by DExTeR as the teacher model effectively translate into better performance for student detectors, regardless of the specific student model used.

\begin{figure*}[h!]
\centering
\includegraphics[width=\textwidth]{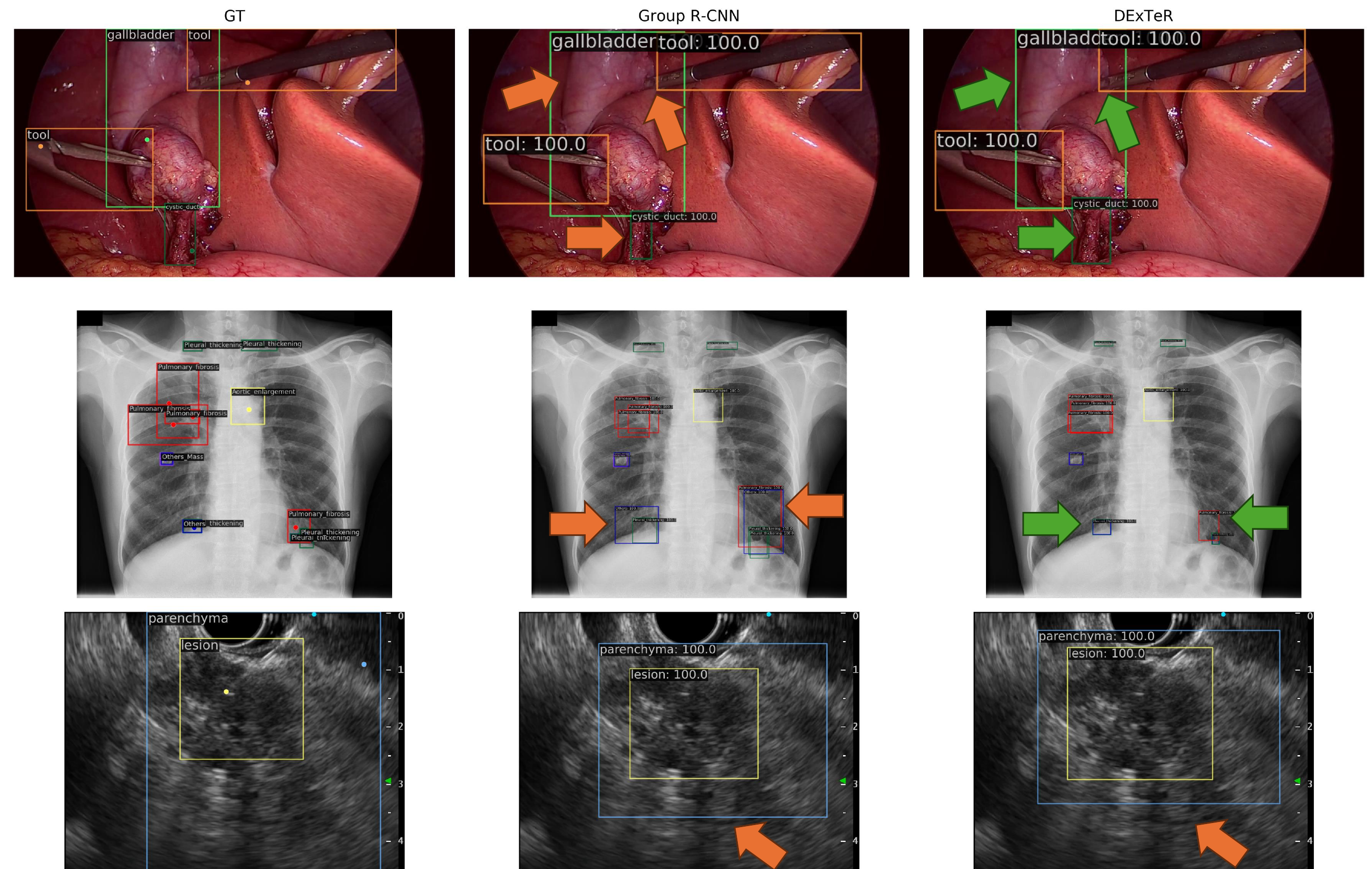}
\caption{Qualitative results for Endoscapes, VinDr-CXR, and EUS-D130 datasets (rows). The first column shows the ground truth bounding boxes and point prompts. The second column presents predictions from Group R-CNN, and the third column shows predictions from DExTeR.}
\label{fig:quali}
\end{figure*}

\subsection{Ablation Studies}

To demonstrate the impact of our proposed components, we conduct ablation studies on the Endoscapes dataset with subsets of $12.5\%$ and $50\%$ well-labeled samples. We limit interference by fixing point annotations on the test set, focusing solely on the efficacy of the Point-to-Box regressor, rather than training a full object detector with pseudo-labeled images.
\\
\\
\noindent \textbf{CLICK Experts Components}. To investigate the impact of each expert within our CLICK-MoE module on the Endoscapes dataset, we systematically activate and deactivate combinations of Class, Instance, and Common Knowledge (CK) experts. This method allows us to assess their individual and collective contributions to model performance, as detailed in Table~\ref{tab:ablation_click_moe}. Note that using the CK expert alone is equivalent to the vanilla design of the Point-DETR decoder. Utilizing all three expert types (i.e. our proposed CLICK-MoE) results in the highest mean Average Precision (mAP) scores, with 27.8\% and 33.9\% on 12.5\% and 50\% labeled samples, respectively. In contrast, employing only the CK expert yields lower mAPs of 24.0\% and 31.4\%. Using the Class expert alone results in significantly poorer outcomes, underscoring its limited effectiveness in isolation. However, pairing the CK expert with either the Class or Instance experts markedly improves performance, indicating that the specialized insights provided by the Class and Instance experts effectively complement the broad perspectives offered by the CK expert, thereby enhancing the model’s accuracy and robustness. This decoupled and structured approach highlights not only the individual value of each expert type but also their synergistic benefits.
\\

\begin{table}[h!]
\begin{threeparttable}
\caption{Impact of Common Knowledge (CK), Class, and Instance Experts on CLICK-MoE performance using the Endoscapes dataset, reporting mean Average Precision (mAP \%) for Point-to-Box models.}
\label{tab:ablation_click_moe}
\begin{tabular*}{\columnwidth}{@{\extracolsep{\fill}}ccccccc}
\toprule

\multirow{2}{*}{CK} & \multirow{2}{*}{Class} & \multirow{2}{*}{Inst.} & \multicolumn{2}{c}{$12.5\%$} & \multicolumn{2}{c}{$50.0\%$} \\
\cmidrule(lr){4-5} \cmidrule(lr){6-7}
 & & & mAP & mAP@50 & mAP & mAP@50 \\
\midrule

\cmark & & & $24.0$ & $43.6$ & $31.4$ & $54.3$
\\
& \cmark & & $12.2$ & $33.0$ & $28.4$ & $52.7$
\\
& & \cmark & $22.9$ & $39.1$ & $30.3$ & $53.9$
\\
\midrule
\cmark & \cmark & & $26.3$ & $46.5$ & $32.8$ & $56.5$
\\
\cmark & & \cmark & $25.1$ & $46.3$ & $32.3$ & $56.0$
\\
& \cmark & \cmark & $23.0$ & $39.1$ & $31.7$ & $55.2$
\\
\midrule
\cmark & \cmark & \cmark & $\textbf{27.8}$ & $\textbf{47.9}$ & $\textbf{33.9}$ & $\textbf{57.4}$
\\

\bottomrule
\end{tabular*}
\end{threeparttable}
\end{table}


\noindent \textbf{Comparison with vanilla MoE}. As a follow-up to our initial experiment, we aim to evaluate the efficacy of our proposed CLICK-MoE by comparing it against a standard Mixture of Experts (MoE) incorporation in the decoder blocks. This comparison is designed to determine the necessity of our design and to discern whether the observed performance gains are attributable to the increased number of parameters or to the specific architecture of CLICK-MoE. We employ a sparse MoE configuration, as detailed in \cite{MoE_review}, which utilizes a router network to select the top-2 experts from the available pool. Table~\ref{tab:ablation_MoE} presents the comparative results for configurations with 3, 5, and 8 experts, alongside our CLICK-MoE and the classic FFN.

We observe that replacing the traditional FFN with a sparse MoE configuration results in only modest performance gains, with an increase of +0.5\% and +0.3\% mAP for 12.5\% and 50\% labeled samples, respectively. These minimal improvements underscore that the significant performance gains of our approach are not solely due to the use of an MoE framework but from the explicit integration of CLass, Instance, and Common Knowledge experts, demonstrating the effectiveness of our CLICK-MoE design.
\\

\begin{table}[h!]
\begin{threeparttable}
\caption{Comparison of refinement layer types in the decoder blocks: We compare the FFN, Mixture of Experts (MoE) with varying numbers of experts, and our proposed CLICK-MoE on the Endoscapes dataset, reporting mean Average Precision (mAP \%) for Point-to-Box models.}

\label{tab:ablation_MoE}
\begin{tabular*}{\columnwidth}{@{\extracolsep{\fill}}ccccc}
\toprule

\multirow{2}{*}{Layer Type} & \multicolumn{2}{c}{$12.5\%$} & \multicolumn{2}{c}{$50.0\%$}
\\
\cmidrule(lr){2-3} \cmidrule(lr){4-5}

& mAP & mAP@50 & mAP & mAP@50 \\
\midrule

FFN
& $24.0$ & $43.6$ & $31.4$ & $54.3$\\
MoE (3)
& $24.1$ & $43.6$ & $31.5$ & $54.2$\\
MoE (5)
& $24.5$ & $43.7$ & $31.7$ & $54.9$\\
MoE (8)
& $24.2$ & $43.7$ & $31.7$ & $54.8$\\
CLICK-MoE
& $\mathbf{27.8}$ & $\mathbf{47.8}$ & $\mathbf{33.9}$ & $\mathbf{57.4}$\\

\bottomrule
\end{tabular*}
\end{threeparttable}
\end{table}

\noindent \textbf{Number of Points in Multi-point Training Strategy}. In our Multi-point training strategy, we select $N$ points per instance to form $N$ query groups, which are decoded in parallel in a group-wise manner. This approach encourages the network to accurately predict box instances irrespective of the point location on the instance, enhancing overall supervision quality. To evaluate the influence of $N$ on model performance, we systematically vary the number of query groups and assess the impact on Point-to-Box regression.

As shown in Table~\ref{tab:ablation_n_points}, performance significantly improves when multiple points are sampled during training. We observe gradual improvements from using $1$ to $8$ points. However, performance starts to saturate and even decrease with \(N \geq 10\), as the model begins to overfit. The increased supervision from more points, while initially beneficial, becomes counterproductive beyond a certain threshold. Based on these results, we choose \(N = 8\) as the default setting in our DExTeR model.\\

\begin{table}[h!]
\begin{threeparttable}
\caption{Comparison of number of points used in our multi-point training strategy on the Endoscapes dataset, reporting mean Average Precision (mAP \%) for Point-to-Box models.}
\label{tab:ablation_n_points}

\begin{tabular*}{\columnwidth}{@{\extracolsep{\fill}}ccccc}
\toprule

\multirow{2}{*}{\# points} & \multicolumn{2}{c}{$12.5\%$} & \multicolumn{2}{c}{$50.0\%$}
\\
\cmidrule(lr){2-3} \cmidrule(lr){4-5}

& mAP & mAP@50 & mAP & mAP@50 \\
\midrule

$1$
& $27.8$ & $47.8$ & $33.9$ & $57.4$\\
$2$
& $28.7$ & $49.1$ & $34.1$ & $58.7$\\
$4$
& $30.1$ & $51.2$ & $35.6$ & $61.2$\\
$6$
& $31.1.$ & $52.3$ & $36.4$ & $62.8$\\
$8$
& $\mathbf{31.4}$ & $\mathbf{52.8}$ & $\mathbf{36.8}$ & $\mathbf{62.9}$\\
$10$
& $31.4$ & $52.5$ & $36.7$ & $62.9$\\
$12$
& $31.0$ & $52.2$ & $36.5$ & $61.7$\\

\bottomrule
\end{tabular*}
\end{threeparttable}

\end{table}

\noindent \textbf{Class-guided Multi-Scale Deformable Attention}.
To emphasize the importance of Class-guided MSDA, we compare it with an alternative implementation that excludes the class embedding during MSDA computation, adding it only afterward (i.e. vanilla MSDA). As shown in Table~\ref{tab:ablation_class_guided}, this modification leads to degraded performance, demonstrating that class information is essential for accurately predicting deformable attention offsets.

\begin{table}[h!]
\begin{threeparttable}
\caption{Comparison of MSDA and class-guided MSDA. A modified version of Deformable Point-DETR is created by removing the class embedding during the computation of prompt-to-vision token deformable attention, with the class embedding added only after the MSDA step to evaluate performance without class-specific guidance during attention computation.}
\label{tab:ablation_class_guided}

\begin{tabular*}{\columnwidth}{@{\extracolsep{\fill}}ccccc}
\toprule

\multirow{2}{*}{class-guided Att.} & \multicolumn{2}{c}{$12.5\%$} & \multicolumn{2}{c}{$50.0\%$}
\\
\cmidrule(lr){2-3} \cmidrule(lr){4-5}

& mAP & mAP@50 & mAP & mAP@50 \\
\midrule

$ $
& $22.4$ & $42.7$ & $30.8$ & $53.5$\\
\cmark
& $\textbf{24.0}$ & $\textbf{43.6}$ & $\textbf{31.4}$ & $\textbf{54.3}$\\

\bottomrule
\end{tabular*}
\end{threeparttable}

\end{table}

%% file: medIA/main/06-conclusion.tex
\section{Conclusion}
In this work, we introduce DExTeR, an advanced Point-to-Box regressor built on the foundation of Point-DETR, specifically designed for Weakly Semi-Supervised Object Detection. DExTeR incorporates several key innovations: Class-aware Multi-Scale Deformable Attention for more effective attention sampling, a Class, Instance, and Common Knowledge Mixture of Experts (CLICK-MoE) for precise query refinement, and a multi-point training strategy to enhance training efficiency and robustness. These enhancements enable DExTeR to surpass prior methods on the Endoscapes, VinDr-CXR, and EUS-D130 datasets, demonstrating its adaptability and effectiveness, particularly in settings with limited high-quality annotations. Furthermore, we show that our approach translates into improved performance for downstream student models. We believe that DExTeR provides a significant step forward in reducing dependency on fine-grained annotations, offering a scalable and efficient solution for object detection in challenging domains such as medical imaging.

%% file: medIA/main/07-acknowledgement.tex
\section*{Acknowledgment }
This work was supported by French state funds managed within the ’Plan Investissements d’Avenir’ funded by the ANR under references ANR-21-RHUS-0001 (DELIVER) and ANR-10-IAHU-02 (IHU Strasbourg). This work was performed using HPC resources managed by CAMMA, IHU Strasbourg, Unistra Mesocentre, and GENCI-IDRIS (Grant AD011013710R2).

\section*{Declaration }
During the preparation of this work, the authors used ChatGPT for minor parts of the manuscript. All AI-generated text was reviewed and edited by the authors, who take full responsibility for the final content.